\documentclass[11pt]{article}
\usepackage[final]{acl}

\usepackage{times}
\usepackage{latexsym}

\usepackage[T1]{fontenc}

\usepackage[utf8]{inputenc}

\usepackage{microtype}

\usepackage{inconsolata}

\usepackage{enumitem}

\usepackage{graphicx}
\usepackage{pgfplots}
\pgfplotsset{width=\textwidth,compat=1.9}

\usepackage{tabu}
\usepackage{booktabs}
\usepackage{multirow}
\usepackage{multicol}
\usepackage{float}
\usepackage{makecell}
\usepackage{tabularx}

\usepackage{amsmath}
\usepackage{amssymb}
\usepackage{amsfonts}
\usepackage{amsthm}
\usepackage{cancel}
\usepackage{bbding}

\usepackage{algorithm,algorithmic}

\usepackage{url}

\DeclareMathOperator*{\THR}{THR}

\definecolor{red}{RGB}{254,76,97}
\definecolor{orange}{RGB}{243,156,17}
\definecolor{yellow}{RGB}{255,193,22}
\definecolor{green}{RGB}{82,196,26}
\definecolor{cyan}{RGB}{52,152,219}
\definecolor{blue}{RGB}{40,120,181}
\definecolor{purple}{RGB}{157,61,207}
\definecolor{black}{RGB}{14,29,105}
\definecolor{gray}{RGB}{191,191,191}
\definecolor{DDDD}{RGB}{0,0,0}
\newcommand{\wyh}[1]{\textcolor{DDDD}{#1}}

\newcommand{\whl}[1]{\textcolor{DDDD}{#1}}
\newcommand{\rjf}[1]{\textcolor{DDDD}{#1}}

\newcommand{\wyhnew}[1]{\textcolor{DDDD}{#1}}
\newcommand{\wyha}[1]{\textcolor{DDDD}{#1}}
\newcommand{\rjfnew}[1]{\textcolor{DDDD}{#1}}

\title{TinyR1-32B-Preview: Boosting Accuracy with Branch-Merge Distillation}

\author{
\textbf{Lin Sun}\textsuperscript{1},
\textbf{Guangxiang Zhao}\textsuperscript{1},
\textbf{Xiaoqi Jian}\textsuperscript{1},
\textbf{Yuhan Wu}\textsuperscript{2},
\textbf{Weihong Lin}\textsuperscript{1},
\textbf{Yongfu Zhu}\textsuperscript{1}, \\
\textbf{Qilong Shi}\textsuperscript{2},
\textbf{Aomufei Yuan}\textsuperscript{2},
\textbf{Yuxuan Tian}\textsuperscript{2},
\textbf{Change Jia}\textsuperscript{1},
\textbf{Linglin Zhang}\textsuperscript{1},\\
\textbf{Jinzhu Wu}\textsuperscript{1},
\textbf{Junfeng Ran}\textsuperscript{2},
\textbf{Sai-er Hu}\textsuperscript{1},
\textbf{Zihan Jiang}\textsuperscript{2},
\textbf{Junting Zhou}\textsuperscript{2},\\
\textbf{Wenrui Liu}\textsuperscript{2},
\textbf{Xusen Xiao}\textsuperscript{2},
\textbf{Bin Cui}\textsuperscript{2},
\textbf{Tong Yang}\textsuperscript{2}
\textbf{Xiangzheng Zhang}\textsuperscript{1},
\\
\textsuperscript{1}\,Qiyuan Tech \quad
\textsuperscript{2}\,Peking University
}

\newcommand{\CodeURL}{\url{https://github.com/Qihoo360/TinyR1-32B-Preview}}
\newcommand{\ModelURL}{\url{https://huggingface.co/qihoo360/TinyR1-32B-Preview}}
  
\begin{document}
\maketitle

\begin{abstract}

It is beneficial but challenging to reduce the size of Large Language Models (LLMs) while maintaining their performance. 
%
Existing methods, such as ordinary model distillation, often fail to achieve high accuracy.
%
To address this limitation, we introduce our Branch-Merge distillation approach:
%
First, a student model is \textit{selectively distilled} to yield different experts with respective domain-specific knowledge from a large teacher model;
%
then, we merge these distilled experts in order to build a generalized model with cross-domain knowledge. 
%
%
With our distillation approach, we create TinyR1-32B-Preview, which outperforms the original student across multiple benchmarks, including Mathematics (+5.5), Coding (+4.4) and Science (+2.9), 
%
and achieves comparable performance to DeepSeek-R1 on AIME 2024. 
%
Our Branch-Merge distillation provides a novel solution for creating smaller, high-performing LLMs with reduced computational cost and time. Model are publicly available at\ModelURL{}.

\end{abstract}
\begin{figure*}[htbp]
\centering
\includegraphics[width=0.96\linewidth]{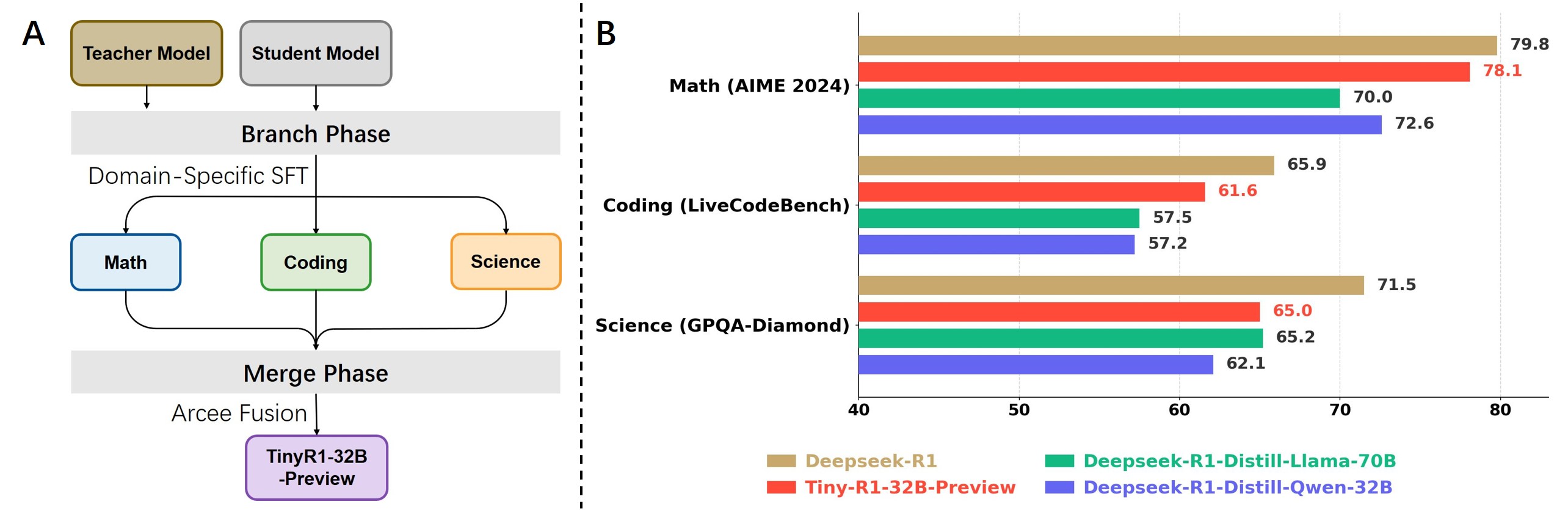}
\caption{(A) A simplified diagram of our Branch-Merge distillation approach. (1) In the Branch phase, each copy of the Initial Model (backbone) is trained on knowledge from a different domain; (2) In the Merge phase, models are merged based on Arcee Fusion rules. (B) Performance Comparison of different LLM models~\cite{mustar2025diagram}. TinyR1-32B-Preview outperforms distilled models of the same size in science, math, and coding and achieves comparable results to Deepseek R1. LiveCodeBench here refers to the 24.08-25.02 subset of full LiveCodeBench.}
\label{fig: TinyR1-main}
\end{figure*}

\section{Introduction}
\label{sec: intro}

\subsection{Motivation}



Recently, DeepSeek-R1 has achieved strong results, and its R1-Distill models \cite{deepseekai2025deepseekr1incentivizingreasoningcapability} show that small models distilled from large teachers can excel at reasoning. Smaller models are also easier to deploy and cheaper at inference. A key challenge is therefore to \emph{shrink model size while maintaining performance}.

In practice, building a small yet capable model typically starts from a pretrained base model and goes through multiple stages, among which Supervised Fine-Tuning (SFT) and Reinforcement Learning (RL) are critical. While RL is widely used to improve reasoning, a strong SFT cold start is often essential to unlock initial capabilities and support further RL gains \cite{qian2025userrltraininginteractiveusercentric}. To our knowledge, the most effective SFT strategy is multi-domain distillation from a larger teacher~\cite{tinybert, deepseekai2025deepseekr1incentivizingreasoningcapability,2025openr1,muennighoff2025s1simpletesttimescaling}. However, naive mixed-data distillation requires careful domain selection and proportion tuning, which is time-consuming and error-prone~\cite{guo-etal-2019-autosem, ji2024pkusaferlhfmultilevelsafetyalignment}. Moreover, optimizing many domains jointly can cause task interference and conflicting gradients, slowing or degrading learning \cite{yu2020gradientsurgerymultitasklearning, jiang2024hummerlimitedcompetitivepreference}. As a result, distilled models may underperform on specialized tasks, even when those tasks appear in training.

To address these issues, we propose a \textbf{Branch-Merge} approach that integrates model merging into distillation. As shown in Figure~\ref{fig: TinyR1-main} (A), it has two phases: (1) \textbf{Branch Phase}, where \wyh{knowledge is selectively distilled from a unified large teacher model (\textit{e.g.}, DeepSeek\rjf{-R1} 671B) into several specialized student models (\textit{e.g.}, \rjf{math}, coding, science) via domain-specific SFT.} (2) \textbf{Merge Phase}, where specialized models are combined into a single unified model, enabling cross-domain transfer while preserving specialized capabilities.





\subsection{Contribution}
\begin{itemize}[leftmargin=*]
\item \textbf{Accuracy}: \wyh{We can see from Figure~\ref{fig: TinyR1-main} (B) that the Branch-Merge distillation approach significantly \rjf{improves} model accuracy. Our distilled Qwen-32B model Tiny-R1-32B-Preview surpasses DeepSeek-R1-Distill-Qwen-32B with about $5\%$ more accuracy, and its math accuracy approaches that of the original R1 teacher, which traditional distillation methods have not yet achieved.} We demonstrate the significant potential of SFT in model tuning, thereby providing a robust starting point for future optimization methods such as RL.

\item \textbf{Simplicity \& Low Cost}: Branch-Merge distillation approach \rjf{significantly} reduces the time and computational costs of the \rjf{merging} stage. Compared to traditional methods, we save 90\% of the time in the \rjf{merging} phase (0.5 hours with 4 H800 GPUs vs. 23 hours with 32 H800 GPUs for \rjf{merged} data retraining). 
The ideal reproduction cost for TinyR1-32B-Preview is 744 H800 GPU hours, approximately \$1500 (excluding ablation experiments and parameter search). 

\item \textbf{Openness}: \wyh{We have publicly released our model at\ModelURL{}. We have also open-sourced the training data, training scripts, model merging code, and evaluation code at\CodeURL{}, ensuring that all core components of TinyR1-32B-Preview are fully public and reproducible.}

\end{itemize}

\section{The Branch-Merge Distillation Approach}
\label{sec: fenhe}

This section describes our branch-merge distillation approach (as shown in Figure \ref{fig: TinyR1-main} (A)), which consists of two phases: Branch and Merge. This two-phase distillation strategy directly addresses the issues of data selection and gradient conflict by decoupling training domains (Branch) and then reconciling them (Merge). Each phase is described in detail below.


\subsection{The Branch Phase}
\label{subsec: fen}

In the Branch phase, we first constructed separate datasets for math, science, and coding. Then, we fine-tuned DeepSeek-R1-Distill-Qwen-32B on each dataset using SFT, resulting in three specialized expert models.

\begin{itemize}[leftmargin=*]
\item \textbf{Math}: We sift 58k samples from 94k questions in NuminaMath1.5 \cite{numina_math_datasets} with corresponding solutions from OpenR1 \cite{2025openr1} trajectories. \wyhnew{The selection is based on three aspects: \textit{question\_type}, \textit{source}, and \textit{correctness\_math\_verify}. 
Ultimately, we adopted a minimal dataset while maintaining comparable performance.}

\item \textbf{Coding}: \wyhnew{The OpenThoughts \cite{openthoughts} dataset is filtered to form 20k trajectories of coding solutions.}

\item \textbf{Science}: \wyhnew{DeepSeek-R1 generates 1 CoT trajectory for each of the 8.6k seed examples (2.7k from the science and health science subsets of \textit{data\_ablation\_full59k} in S1~\cite{muennighoff2025s1simpletesttimescaling}, 1.0k from S1k \cite{muennighoff2025s1simpletesttimescaling}, 4.9k from the science subset of OpenThoughts \cite{openthoughts}), resulting in 8.6k CoT trajectories. }
\end{itemize}

\wyhnew{We apply SFT on DeepSeek-R1-Distill-Qwen-32B with the three datasets to obtain three specialized models. }

\subsection{The Merge Phase}
\label{subsec: he}

In the Merge phase, we use Arcee Fusion~\cite{goddard-etal-2024-arcees} to merge models from different domains. 


The core idea of Arcee Fusion is as follows: when merging a new model of identical size and structure into an existing one, parameters pairs at corresponding positions are processed sequentially. For each pair, the decision of which parameter to keep is based on the magnitude of their difference. The greater the difference, the stronger the preference for adopting the parameter from the new model.
In this merging process, we refer to the existing model as the \textbf{Left Model} and the new model as the \textbf{Right Model},
\wyha{with their parameters denoted by $\theta_L, \theta_R \in \mathbb{R}^{M}$, where \(M\) is the number of model parameters, respectively.}

 The merge process is divided into three steps:

\begin{enumerate}[leftmargin=*, label={\arabic*)}]

\item \textbf{Computing Importance Score.}

We define an importance score vector $S_{IS}\in\mathbb{R}^{M}$ to quantify how salient each coordinate-wise update from the Right Model is, relative to the Left Model.
Concretely, for each parameter tensor (e.g., a weight matrix) we flatten its entries into a length-$N$ vector and treat it as a sample $X=\{x_1,\ldots,x_N\}$ from the Left Model, and similarly $Y=\{y_1,\ldots,y_N\}$ from the Right Model.
We then normalize them into discrete distributions via a softmax transform:
\[
\tilde{x}_{i} = \frac{e^{x_i}}{\sum_{j=1}^N e^{x_j}}+\epsilon,\quad
\tilde{y}_{i} = \frac{e^{y_i}}{\sum_{j=1}^N e^{y_j}}+\epsilon,
\]
where $\epsilon$ is a small constant with empirical default value $10^{-8}$.
We compute a tensor-level divergence $D_{KL}(\tilde{X}\parallel\tilde{Y})$ and broadcast it to obtain coordinate-wise saliency by scaling the parameter delta:
\begin{equation}
S_{IS} = D_{KL}(\tilde{X} \parallel \tilde{Y}) \cdot (\theta_{L}-\theta_{R}).
\end{equation}
Intuitively, $D_{KL}(\tilde{X}\parallel\tilde{Y})$ captures how strongly the Right Model departs from the Left Model within the tensor, while $(\theta_{L}-\theta_{R})$ localizes this departure at each coordinate.

\item \textbf{Calculating Dynamic Selection:}







We sorted $S_{IS}$ and define these variables: $\mathcal{Q}_\frac{1}{4}, \mathcal{Q}_{Med}, \mathcal{Q}_\frac{3}{4}$, which denotes the 25\%, 50\%, 75\% quantile value of the sorted $S_{IS}$, respectively. Additionally, we calculate the interquartile range \(\mathcal{Q}_{IR}\) as \(\mathcal{Q}_\frac{3}{4}- \mathcal{Q}_\frac{1}{4}\). Note that \(\mathcal{Q}_\frac{3}{4}\geq  \mathcal{Q}_\frac{1}{4}\) holds, ensuring \(\mathcal{Q}_{IR}\geq 0\).






\wyha{$\mathcal{S}_{THR} \in \mathbb{R}$ is then defined} via the median  $\mathcal{Q}_{Med}$ and interquartile range $\mathcal{Q}_{IR}$ :

\begin{equation}
    \mathcal{S}_{THR} = \mathcal{Q}_{Med} + \lambda \cdot \mathcal{Q}_{IR},
\end{equation}

where $\lambda$ is a hyperparameter defined to balance coefficient.

Since $\mathcal{S}_{THR}$ is derived from IS statistics, it dynamically adapts to model parameters, ensuring an optimal update ratio.

\item \textbf{Selective Integration}: 

\wyha{The Merged Model is defined as $\theta_{M} \in \mathbb{R}^{M}$.} We applied the following rule to merge $\theta_{L}$ and $\theta_{R}$:

\begin{equation}
    \theta_{M}^{i} =\theta_{L}^i + (\theta_{R}^i-\theta_{L}^i) \cdot \lceil \max(0,\mathcal{S}_{IS}^i - \mathcal{S}_{THR})\rceil.
\end{equation}

Only Right Model's parameters with an importance score above the threshold are retained; otherwise, Left Model's parameters are kept.

\end{enumerate}

\wyhnew{By focusing on the most significant changes, Arcee Fusion avoids over-updating and maintains model stability.} The details of our merge sequence is in Section~\ref{subsec: comparison}.

\paragraph{Why branch-merge helps.}
A data-mixture baseline jointly optimizes a single parameter vector over multiple domains and can suffer from cross-domain gradient interference, which empirically appears as a seesaw effect across domain scores. Our branch-merge strategy first decouples these conflicts by training domain experts independently, and then selectively integrates only high-importance expert updates via Arcee Fusion, approximating the union of expert capabilities rather than a compromise solution. We provide a more detailed theoretical discussion in Appendix~\ref{app:theory_branch_merge}.

\begin{table*}[ht]
\centering
\small
\begin{tabular}{lcccc}
\toprule
\multirow{2}{*}{Model} &  Math & Coding & Science \\
  &  (AIME 2024) & (LiveCodeBench 24.08-25.02) & (GPQA-Diamond) \\
\midrule
DeepSeek-R1-Distill-Qwen-32B\(^{\dagger}\) & 72.6 (9.6k Tokens)  & 57.2 (10.1k Tokens) & 62.1 (5.3k Tokens) \\
DeepSeek-R1-Distill-Llama-70B\(^{\dagger}\) &  70.0 & 57.5 & 65.2 \\
DeepSeek-R1\(^{\dagger}\) & 79.8 (9.6k Tokens) & 65.9 (10.4k Tokens) & 71.5 (5.3k Tokens) \\
\midrule
TinyR1-32B-Preview (Ours) & 78.1  (11.8k Tokens) & 61.6 (12.4k Tokens) & 65.0 (8.6k Tokens) \\
\bottomrule
\end{tabular}
\caption{Performance comparison on benchmark datasets. All scores are reported as pass@1. Scores reported from DeepSeek-R1 paper \cite{deepseekai2025deepseekr1incentivizingreasoningcapability} are noted with \(^{\dagger}\). The number in parentheses represents the average output token length (including the chain of thought), obtained from our testing.}
\label{tab: main_results}
\vspace{-0.3cm}
\end{table*}

\begin{table*}[ht]
\centering
\small
\begin{tabular}{lccccc}
\toprule
\multirow{2}{*}{Model}  & Math & Coding & Science & Merging Time \\
  & (AIME 2024) & (LiveCodeBench) & (GPQA-Diamond) & (GPU Hours) \\
\midrule
Math Expert  & 73.1 & - & - &-  \\
Coding Expert  & - & 63.4 & - &-  \\
Science Expert  & - & - & 64.5 &- \\
\midrule
Data Mixture  & 75.3 & 61.0 & \textbf{65.7} & 740 h \\
\midrule
Merging: (Math \& Coding) \& Science & 77.3 & \textbf{63.8} & 64.0 & \textbf{4h} \\
Merging: (Math \& Science) \& Coding & \textbf{78.1} & 61.6 & 65.0 & \textbf{4h} \\
\midrule
TinyR1-32B-Preview & \textbf{78.1} & 61.6 & 65.0 & \textbf{4h} \\
\bottomrule
\end{tabular}
\caption{Performance comparison between backbone experts, the data-mixture model, and merged model. All scores are reported as pass@1. LiveCodeBench here refers to the 24.08-25.02 subset of full LiveCodeBench.}
\label{tab: merging_results}
\vspace{-0.3cm}
\end{table*}

\section{Experiment}
\label{sec: experiment}

We choose Deepseek-R1 \cite{deepseekai2025deepseekr1incentivizingreasoningcapability} and its distilled DeepSeek-R1-Distill-Qwen-32B and DeepSeek-R1-Distill-Llama-70B as baselines. 
Additionally, we conducted a comprehensive ablation study. We compared TinyR1-32B-Preview with: (a) three domain expert models (Math Expert, Coding Expert, Science Expert) before merging; (b) a ‘Data Mixture’ model trained on a combined Math/Coding/Science dataset; and (c) variants of our model obtained via different merging sequences. Detailed experiment setup is discussed in Appendix~\ref{subsec: setup}.

\subsection{Main Results}

\wyhnew{We compare our TinyR1-32B-Preview model and other models in Table \ref{tab: main_results}.}

\begin{itemize}[left=0pt]
    \item \wyhnew{In terms of accuracy, we outperform our backbone model, DeepSeek-R1-Distill-Qwen-32B (Math +5.5, Coding +4.4, Science +2.9), and generally surpass DeepSeek-R1-Distill-Llama-70B (Math +8.1, Coding +4.1, Science -0.2), approaching the performance of DeepSeek-R1 (Math -1.7, Coding -4.3, Science -6.5).}
    \item \wyhnew{In terms of inference cost, our model generates slightly more output tokens than DeepSeek-R1 (Math +23\%, Coding +19\%, Science +62\%).} Note that our model is smaller compared to DeepSeek-R1, making it more suitable for local deployment by users and small groups.
\end{itemize}

\subsection{Ablation Study}
\label{subsec: comparison}

As shown in Table~\ref{tab: merging_results}, we made a comprehensive ablation study to explore if our merging distillation approach works.

Compared to the domain-specific experts, the Data Mixture model surpasses them in math and science but shows decreased performance in coding. This is a seesaw problem that traditional data mixture is difficult to solve \cite{zamir2018taskonomy, liu2019meta, radford2021improving}.
In comparison, our merged model improves performance in math and science domains while largely retaining the coding capability. 
We also compared two different model-merging sequences: (1) first merging Math with Coding, then merging with Science; (2) first merging Math with Science, then with Coding. The results show only minor performance differences between these sequences.

In addition, the average Merging time (GPU hours) of the current Data Mixture is 740 hours\footnote{Note that the merging time is the SFT experiment time, excluding the SFT time for single experts and the downstream evaluation time of the Data Mixture model}. On the contrary, the average merging time of Tiny-R1-32B-preview is 4 hours. In summary, we only used 0.5\% of the Data-Mixture computational overhead on merging models to surpass the effect of traditional data mixture methods. 
\wyhnew{In addition to reducing computational overhead, model merging significantly accelerates our model release process by avoiding the delays introduced by mixed-data re-SFT on the development model.
The model merging is a ``free-lunch'' approach, as it reduces costs and increases efficiency at the same time.} The further studies are in Appendix \ref{sec:furthur:ablation}.

\section{Conclusion}
\label{sec: conclusion}

We introduce TinyR1-32B-Preview, a model using the Branch-Merge distillation approach to boost reasoning accuracy while preserving efficiency. 
We achieve significantly higher accuracy than our backbone model, DeepSeek-R1-Distill-Qwen-32B, and generally outperform DeepSeek-R1-Distill-Llama-70B while approaching the performance of DeepSeek-R1. Moreover, its smaller parameter size makes it more efficient and better suited for local deployment by users and small groups. For future optimization methods like RL, our model provides a robust starting point.

\section*{Limitations}

As a small-scale research project, we have the following limitations:
\begin{itemize}[leftmargin=*]
   \item \textbf{Incomplete Benchmark Evaluations} – Our report does not include many reasoning benchmarks such as Codeforces rating and SWE-bench. 
    \item \textbf{Unfinished Future Works} - There remains several promising directions for extending the training framework, such as enhancing its instruction-following (IF) and safety capabilities. Additionally, our approach can be integrated with reinforcement learning (RL) to further improve training outcomes, and we look forward to validating this potential in future studies.
\end{itemize}

\bibliography{anthology,custom}

\appendix

\section{Experiment Setup Details}
\label{subsec: setup}

\subsection{Training Details}

We employ DeepSeek-R1-Distill-Qwen-32B as our backbone model.
Leveraging the 360-Llama-Factory \cite{360-llama-factory} training framework, we develop three domain-specific expert models applying 16384 sequence length with constructed Math, Coding, and Science datasets.
\begin{itemize}[leftmargin=*]
\item Math Expert: The math expert model is trained with 5 epochs, batch size 96, and the learning rate is set to constant 1e-5.

\item Science Expert: The science expert model is trained with 5 epochs, batch size 32 with the neat packing mechanism \cite{tay2020efficient, henry2019memory, dean2018scaling}, and the learning rate is set to cosine 1e-5.

\item Coding Expert: The coding expert model is trained with 15 epochs, batch size 96 with the neat packing mechanism, and the learning rate is set to constant 1e-5.
\end{itemize}


We merged the models trained separately in three fields into a single model. We use the Arcee merging \cite{goddard-etal-2024-arcees} method with the \(\theta\)=1.5 and threshold $\THR$=0.5. We will compare different model merging methods in Section~\ref{subsec: comparison}.

\subsection{Evaluation Details}

For evaluation, we compare the performance on three benchmark datasets: AIME 2024 for Math, LiveCodeBench (24.08-25.02) for Coding, and GPQA-Diamond for Science. The accuracy is calculated as an average pass@1 across 16, 4, and 4 independent trials for these benchmarks, respectively. Meanwhile, we did not use a greedy way to evaluate the model due to its long-COT output, we set the max tokens to 32768 and evaluated the models with Temperature=0.6 and Top-p=0.95 as recommended in DeepSeek-R1 \cite{deepseekai2025deepseekr1incentivizingreasoningcapability}. We tried various open-source frameworks for the evaluation on livecodebench and ultimately selected the evaluation code from FuseAI \cite{wan2024fusechat} utilizing the vLLM implementation, as it can reproduce the effects of the DeepSeek-R1 and its distilled models.

\section{Related Work}
\label{sec: related_work}

\subsection{Model Distillation}
\label{subsec: model_distillation}

\rjfnew{Knowledge Distillation (KD) \cite{DBLP:RomeroBKCGB14, hinton2015distillingknowledgeneuralnetwork} has been proposed to create cheaper strong models \cite{Gou_2021, hu2023teacherstudentarchitectureknowledgedistillation, yang2024surveyknowledgedistillationlarge, xu2024surveyknowledgedistillationlarge}. Primarily, recognizing the disparities between proprietary and open-source LLMs, KD techniques have surged to bridge the performance gap between these models. Distillation methods can be categorized into two types: (1) the logits-based methods \cite{hinton2015distillingknowledgeneuralnetwork}, which transfer knowledge at the logits level, and (2) the feature-based methods \cite{DBLP:RomeroBKCGB14}, which transmit knowledge through intermediate features.}

\rjfnew{Compared to traditional knowledge distillation techniques \cite{Gou_2021}, data augmentation (DA) \cite{feng-etal-2021-survey} has emerged as an effective method for distilling knowledge in large language models (LLMs). In this approach, a small seed of knowledge is used to prompt LLMs, enabling them to generate additional data tailored to specific domains or skills \cite{alpaca}. More recently, an API-based strategy has gained attention, where open-source LLMs serve as teachers to self-improve through self-distillation \cite{yuan2024selfrewardinglanguagemodels, chen2024selfplayfinetuningconvertsweak}. By applying a range of distillation techniques, this strategy effectively narrows the performance gap between closed-source and open-source models \cite{vicuna2023, xu2023wizardlmempoweringlargelanguage}. In this context, the method involves using only the outputs of the teacher model via an API. This strategy includes approaches such as In-Context Learning \cite{huang2022incontextlearningdistillationtransferring}, Chain-of-Thought \cite{li2022explanationslargelanguagemodels}, and Instruction Following \cite{wang2023selfinstructaligninglanguagemodels}. In specialized fields, like science \cite{zhang2024sciinstructselfreflectiveinstructionannotated}, where domain-specific knowledge and accuracy are essential, distillation allows open-source models to significantly improve their performance by learning from proprietary models that are extensively trained and fine-tuned in these domains.}

\subsection{Model Merging}
\label{subsec: model_merging}

\whl{Recent advances in model merging have explored diverse strategies \cite{ilharco2022editing, yadav2023ties, davari2024model, deep2024della} to combine neural network parameters while preserving or enhancing performance. Early approaches focused on linear interpolation techniques, such as weight averaging (Wortsman et al., 2022), where models finetuned from shared pretrained checkpoints are merged via arithmetic mean. While computationally efficient, these methods assume approximate parameter space alignment and often degrade when models exhibit divergent optimization trajectories \cite{frankle2020linear, izmailov2018averaging, neyshabur2020being, fort2020deep, wortsman2022model, choshen2022fusing, ilharco2022patching}. }

\whl{
Theoretical underpinnings for these methods derive from studies on loss landscape geometry~\cite{ilharco2022editing, li2018visualizing, garipov2018loss, draxler2018essentially, kuditipudi2019explaining, fort2019deep, czarnecki2019deep, wortsman2021learning, benton2021loss, entezari2021role, li2022branch, lubana2023mechanistic}. Research on flat local minima \cite{kaddour2022flat, wortsman2022model,keskar2016large,dziugaite2017computing} dating back from the 1990s \cite{hochreiter1994simplifying, hochreiter1997flat} suggests that averaged weights reside in flatter regions of the loss surface, correlating with improved out-of-distribution generalization. Further analyses \cite{daheim2023model, matena2022merging} formalize model merging as identifying connected basins in parameter space, where interpolated solutions maintain low loss. Empirical validations, such as model soups \cite{wortsman2022model}, corroborate that aggregated weights often outperform individual models, particularly under distribution shifts.
}

\section{Theoretical Discussion}
\label{app:theory_branch_merge}

\subsection{Why data mixture can underperform: gradient interference}
Consider multi-domain training over $\mathcal{D}=\{\text{math},\text{code},\text{sci}\}$. A data-mixture baseline optimizes a single parameter vector $\theta$ to minimize the summed objective
\begin{equation}
\min_{\theta}\ \sum_{d\in\mathcal{D}} \mathbb{E}_{x\sim d}\big[\mathcal{L}_{d}(\theta; x)\big].
\label{eq:mixture_objective}
\end{equation}
Let $g_d(\theta)=\nabla_{\theta}\mathbb{E}_{x\sim d}[\mathcal{L}_{d}(\theta;x)]$ denote the gradient contributed by domain $d$.
A well-studied issue in multi-task learning is \emph{gradient interference}, where gradients from different domains can conflict, e.g.,
\begin{equation}
\langle g_{i}(\theta), g_{j}(\theta)\rangle < 0,\quad i\neq j,
\label{eq:grad_conflict}
\end{equation}
which implies that a descent step that improves one domain may increase the loss of another (see, e.g., discussions around conflicting gradients in multi-task optimization).
Intuitively, the optimizer is pushed toward a compromise region where gradients partially cancel, limiting progress toward any single domain-preferred solution. This provides a principled explanation for the empirically observed seesaw effect in mixture baselines, where gains on one benchmark can coincide with regressions on another.

\subsection{Branch: decoupling optimization landscapes via domain experts}
Our Branch phase trains domain-specific experts independently, producing expert parameters $\theta^{(d)}$ for each domain $d$ from a shared initialization (e.g., the base model).
This decouples optimization landscapes: each expert can move toward a domain-preferred region without being constrained by conflicting gradients from other domains.

Let $\Delta\theta^{(d)}=\theta^{(d)}-\theta_{\text{base}}$ denote the expert update relative to the base model. In the over-parameterized regime typical for LLMs, it is empirically common that effective task updates are sparse and/or concentrated on subsets of coordinates.
A stylized assumption capturing this behavior is that for distinct domains $i\neq j$, the overlap of their dominant update supports is limited, or equivalently the normalized alignment is small:
\begin{equation}
\frac{\langle \Delta\theta^{(i)}, \Delta\theta^{(j)}\rangle}{\|\Delta\theta^{(i)}\|\,\|\Delta\theta^{(j)}\|} \approx 0.
\label{eq:approx_orthogonal}
\end{equation}
Under such a regime, independent experts can encode domain-specific capabilities in partially distinct functional subspaces of the parameter space.

\subsection{Merge: saliency-guided selective integration as a high-pass gate}
The Merge phase aims to combine complementary expert capabilities while avoiding dilution from naïve weight averaging.
We merge an existing model (Left) with a new expert (Right), denoting parameters by $\theta_L,\theta_R\in\mathbb{R}^M$, and the merged model by $\theta_M\in\mathbb{R}^M$.
Arcee Fusion produces a coordinate-wise importance score $S_{IS}\in\mathbb{R}^M$ and a data-adaptive threshold $S_{THR}\in\mathbb{R}$ (computed from robust statistics of $S_{IS}$, such as the median and interquartile range). The merge rule can be written in a gate form:
\begin{equation}
\theta_M^i = \theta_L^i + (\theta_R^i - \theta_L^i)\cdot \mathbb{I}\!\left(S_{IS}^i > S_{THR}\right)
\label{eq:gate_merge}
\end{equation}
Equation~\eqref{eq:gate_merge} makes explicit that Arcee Fusion performs \emph{saliency-guided selective integration}: only coordinates with sufficiently high importance are copied from the Right Model; otherwise the Left coordinate is preserved.

This can be interpreted as a data-driven \emph{high-pass filter} on expert updates. Large, salient parameter shifts---which more likely encode crucial domain-specific logic---are retained, while small or noisy deviations are suppressed. When combined with the Branch phase (which reduces interference and encourages partially distinct expert updates as in Eq.~\eqref{eq:approx_orthogonal}), the gate in Eq.~\eqref{eq:gate_merge} approximates a union of expert capabilities across domains, rather than the compromise solution encouraged by mixture training in Eq.~\eqref{eq:mixture_objective}.
This perspective aligns with our empirical results, where the merged model consistently improves over the data-mixture baseline on domain benchmarks.

\section{Further Study}

\label{sec:furthur:ablation}

\subsection{Comparison of Different Merging Methods}

\begin{figure}[h!]
\centering
\includegraphics[width=0.96\linewidth]{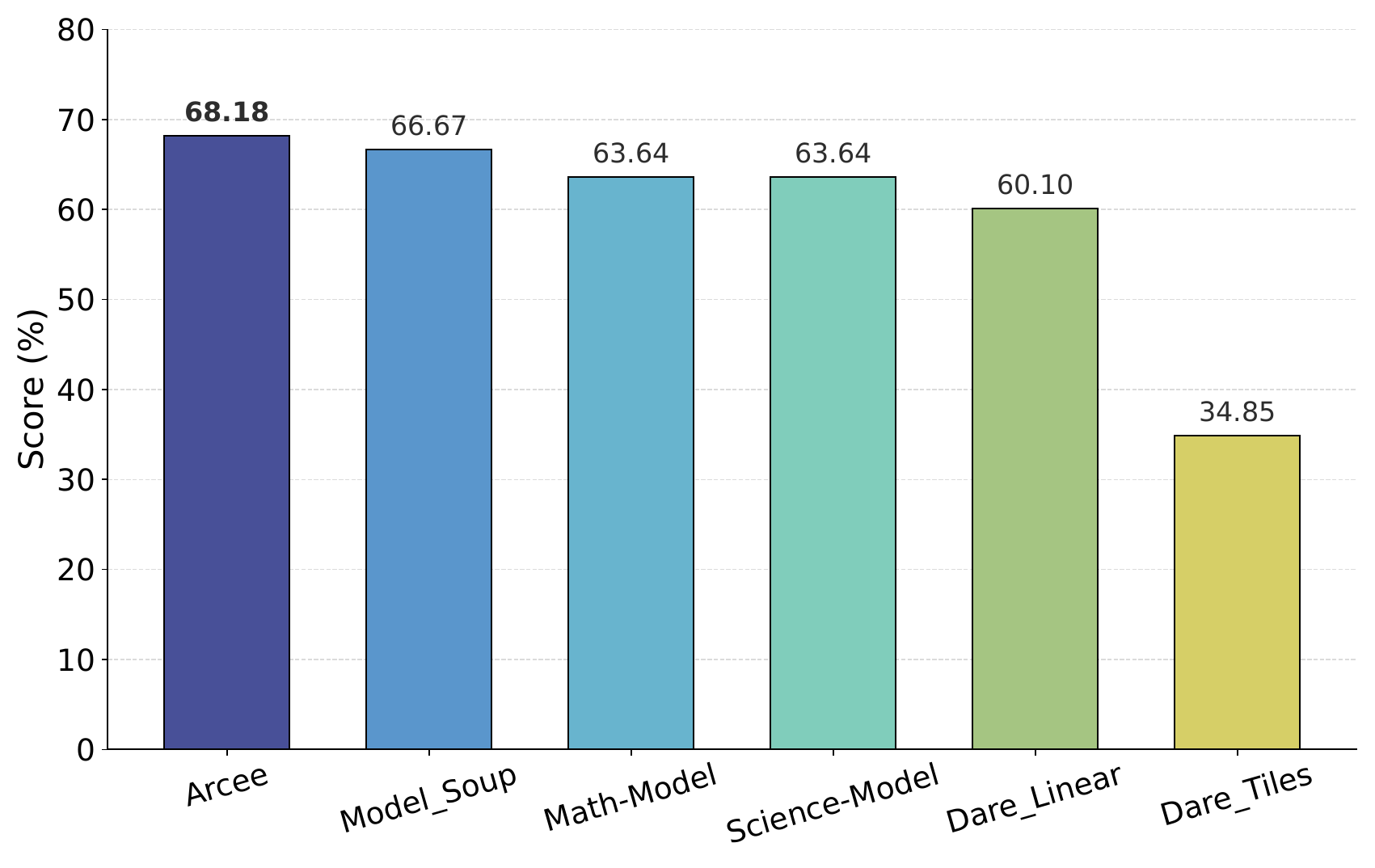}
\caption{Performance Comparison of merged models on the GPQA-Diamond benchmark.}
\label{fig: merging_comparison}
\end{figure}

A comparison with other model merging methods appears in Figure~\ref{fig: merging_comparison}. We compare various merging methods on merging models trained from the math and science domains, and we find that Arcee achieves the highest scores on GPQA-Diamond. We found a similar method ranking on the AIME 2024 benchmark, and we omit the graph.

\subsection{Generalization Beyond Target Benchmarks}

Beyond our primary evaluation suite (AIME, GPQA, and LiveCodeBench), we further assess generalization on three auxiliary benchmarks: IFEval (instruction following), BBH \cite{suzgun2023challenging}, and HumanEval \cite{chen2021evaluating}. Among them, IFEval is the clearest out-of-domain (OOD) probe for our setting, since our distillation data does not contain instruction-following supervision; thus any changes on IFEval reflect OOD generalization. In contrast, BBH and HumanEval overlap more with the broad reasoning/coding skills already exercised by our main benchmarks, so we view them as complementary generalization checks rather than strictly OOD.

\begin{table}[h]
\centering
\small
\begin{tabular}{lccc}
\toprule
Model & IFEval & HumanEval & BBH \\
\midrule
Base Model & 75.5 & 83.5 & 47.0 \\
\midrule
Math Expert & 63.1 & 85.5 & 48.2 \\
Science Expert & 60.4 & 85.5 & 61.1 \\
Code Expert & 62.2 & 85.5 & 63.5 \\
\midrule
TinyR1-32B-Preview & 64.0 & 86.4 & 66.1 \\
\bottomrule
\end{tabular}
\caption{Generalization results beyond our primary benchmarks (AIME, GPQA, LiveCodeBench). IFEval probes instruction-following and serves as an OOD check in our setting; BBH and HumanEval are complementary evaluations of general reasoning and coding.}
\label{tab:generalization_results}
\end{table}

As shown in Table~\ref{tab:generalization_results}, we observe two consistent phenomena on IFEval: (i) models after SFT (and subsequent merging) score lower than the base model (75.5), which is expected and consistent with prior observations \cite{zhou2023instruction, wang2025light}; and (ii) our merged model (TinyR1-32B-Preview) outperforms all individual SFT experts evaluated on IFEval, suggesting that our branch-merge distillation does not further erode—and may partially preserve—instruction-following behavior relative to the pre-merged experts. On HumanEval and BBH, expert models generally improve over the base model, and the merged model achieves the best HumanEval score (86.4) among the evaluated models, indicating that the branch-merge procedure can retain (or enhance) general coding performance.

\end{document}